\documentclass[10pt,twocolumn,letterpaper]{article}

\usepackage{cvpr}
\usepackage{times}
\usepackage{epsfig}
\usepackage{graphicx}
\usepackage{amsmath}
\usepackage{amssymb}

\usepackage{hhline}
\usepackage{array}
\usepackage{booktabs}
\usepackage{multirow}
\usepackage{fancyhdr}
\usepackage[export]{adjustbox}
\usepackage[breaklinks=true,bookmarks=false]{hyperref}

\cvprfinalcopy 

\def\cvprPaperID{****} 

\begin{document}

\title{Empowering Sign Language Communication: Integrating Sentiment and Semantics for Facial Expression Synthesis}

\author{Rafael V. Azevedo\\
UFMG\\
\and
Thiago M. Coutinho\\
UFMG\\
\and
Jo\~ao P. Ferreira\\
UFMG\\
\and
Thiago L. Gomes\\
UFV\\
\and
Erickson R. Nascimento\\
UFMG\\
}

\maketitle
\thispagestyle{fancy}


\begin{abstract}
    
Translating written sentences from oral languages to a sequence of manual and non-manual gestures plays a crucial role in building a more inclusive society for deaf and hard-of-hearing people. Facial expressions (non-manual), in particular, are responsible for encoding the grammar of the sentence to be spoken, applying punctuation, pronouns, or emphasizing signs. These non-manual gestures are closely related to the semantics of the sentence being spoken and also to the utterance of the speaker's emotions. However, most Sign Language Production (SLP) approaches are centered on synthesizing manual gestures and do not focus on modeling the speaker’s expression. This paper introduces a new method focused in synthesizing facial expressions for sign language. Our goal is to improve sign language production by integrating sentiment information in facial expression generation. The approach leverages a sentence's sentiment and semantic features to sample from a meaningful representation space, integrating the bias of the non-manual components into the sign language production process. To evaluate our method, we extend the Frechet gesture distance (FGD) and propose a new metric called Fréchet Expression Distance (FED) and apply an extensive set of metrics to assess the quality of specific regions of the face. The experimental results showed that our method achieved state of the art, being superior to the competitors on How2Sign and PHOENIX14T datasets. Moreover, our architecture is based on a carefully designed graph pyramid that makes it simpler, easier to train, and capable of leveraging emotions to produce facial expressions. Our code and pretrained models will be available at: \url{https://github.com/verlab/empowering-sign-language}.

\end{abstract}
\begin{figure*}[t!]
    \centering
 	\includegraphics[width=0.85\linewidth]{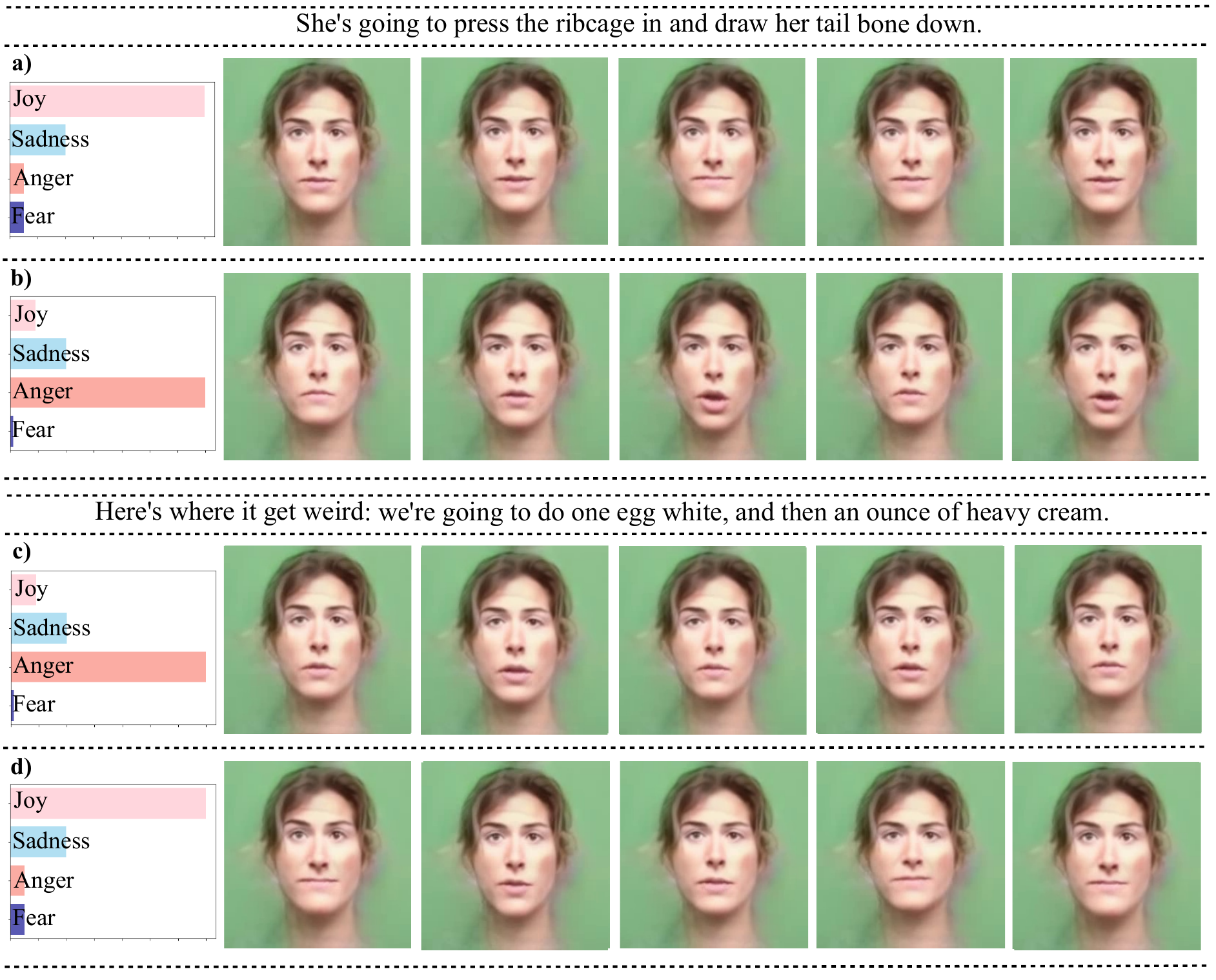}
 	\caption{\textbf{Facial expression synthesis for sign language.} Given a sentence of written text, our method generates distinct non-manual gestures (a, b, c and d) according to the sentiment information used as input to the model. All expressions in the image were automatically generated by our method.}
 	\label{fig:expressions_sentiment_sp8}
\end{figure*}
\section{Introduction}

Sign language is the primary form of communication for the deaf community. Many deaf or hard-of-hearing people first learn to speak in sign language, then learn the vocabulary and grammar of oral languages, such as English, as a second language. Having English as a second language creates many social and communication challenges for the deaf community. For instance, hard-of-hearing are medically undeserved worldwide due to the lack of doctors who can understand and use sign language~\cite{mckee, masuku}. Thus, translating written sentences from oral languages to a sequence of manual and non-manual gestures plays a crucial role in narrowing the gap between oral language speakers and the deaf community. Despite the substantial progress made by sign language recognition~\cite{rajal, rajal2, koller}, synthesizing manual and non-manual gestures from text is still in its prelude phase ~\cite{everybodysign}.



Sign languages are visual and richly composed of manual and non-manual components. While manual components are responsible for coding the hand signals corresponding to the words in the oral language, the non-manual components are responsible for encoding the sentence's grammar, applying punctuation and pronouns, or emphasizing signs. In other words, to properly convey the message when communicating, speakers of sign languages must express it through hand signals, gestures, facial expressions, and body language. For example, in American Sign Language (ASL), facial expressions are closely linked to the semantics of the sentence spoken and the exclamation of the speaker's emotions, as shown in the work of Pfau~\etal~\cite{nonmanuals}. When speaking a sentence containing the sign corresponding to the word {\it happy}, the speaker can make a more open facial expression indicating happiness or, to disambiguate the meaning of two identical signs, the speaker can make a facial gesture to emphasize its meaning by using different parts of the face, such as cheeks or eyebrows. The primary motivation of this work is to ensure the message is conveyed accurately when applying sign language production methods. Sign language relies on various components, including facial expressions and gestures, to communicate the intended message effectively. Therefore, this work aims at improve sign language production by integrating sentiment information in facial expression generation, which is a key issue that has yet to be addressed by previous research.



This paper introduces a new method to synthesize facial expressions for sign language. The approach leverages a sentence's sentiment and semantic features to sample from a meaningful representation space, integrating the bias of the non-manual components into the sign language production process (an alluring example is depicted in Figure~\ref{fig:expressions_sentiment_sp8}). We propose using a two-step approach, which first creates a meaningful representation space using Generative Latent Optimization~\cite{bojanowski2018optimizing} and then, the model learns to sample from our space considering essential aspects in a dialogue such as semantics and sentiment. Moreover, inspired by the work of Yoon~\etal~\cite{Yoon2020Speech}, we present a new quantitative metric called Fréchet Expression Distance (FED) as a tool for evaluating the quality of generated expressions. The experimental results showed that our method holds state-of-the-art results, being superior to the competitors.
%
%
%
%

We can summarise the key contributions of this paper as follows: i) A sentiment-aware method capable of generating facial expressions for Sign Language using only textual data; ii) A designed graph convolutional decoder architecture that can be used in other face-related tasks such as Talking Head Animation; iii) An autoencoder model trained to apply the concept of FID to measure the quality of generated facial expressions for Sign Language Production; iv) Extensive experiments and an ablation study showing the effectiveness of our method in the generation of facial expressions for sign language.





\section{Related Work}

\paragraph{Sign Language Production} The field of Sign Language Production (SLP) aims to produce realistic and continuous sign sequences from spoken language. Early works have proposed to apply avatar-based techniques~\cite{Glauert2006VANESSAA, KARPOUZIS200754, avatar_mcdonald} and statistical approaches~\cite{Kayahan_statistical, Kouremenos_statistical}, both relying on intensive manual work. These works were strongly depend on the creation of avatars, pre-generated sequences and phrase lookup schemes, and rule-based methods carefully designed. Recent advances in deep learning techniques have benefited a myriad of areas, particularly Neural Machine Translation (NMT), which seeks to learn to translate one language to another using neural networks. Encoder-decoder architecture~\cite{enc_dec1, enc_dec2} has been one of the most successful approaches, and recently, the Transformers networks~\cite{transformers} became a key force in modeling non-sequential relationships between sentences of different languages. 

Stoll~\etal~\cite{stoll2018} presented one of the first sign language production models based on NMT and Generative Adversarial Networks (GANs)~\cite{gan}. Their model learns how to map texts to glosses, glosses to skeleton poses by using a look-up table, and to produce a rendered human signer. Despite paving the initial path for the use of NMT for Sign Language Production, the proposed method is not able to learn a smooth mapping between the written text and the sequences of gestures, which produces signs that lacks grammatical syntax. Saunders~\etal~\cite{saunders2020progressive} proposed to produce more smooth and stable outputs using an auto-regressive Transformer model trained in an end-to-end manner. The authors introduced a counter decoding technique to produce sequences of continuous sign language poses with different output sizes. Although their model can produce manual gestures in a relatively stable way, it suffers from problems like error propagation~\cite{error_propagation} due to its auto-regressive nature and regression to the mean~\cite{ginosar, yan2019convolutional}. 
%
To circumvent these problems, recent works are tackling the NMT problem in a non-autoregressive manner. Hwang~\etal~\cite{hwang2021non} proposed NSLP-G, which first learns a latent representation of human poses using a Variational Autoencoder~\cite{VAE} and then use a Transformer to predict sequences from this latent representation. However, the NSLP-G model does not consider target sign lengths and suffers from false decoding initiation~\cite{hwang2022nonautoregressive}. Hwang~\etal~\cite{hwang2022nonautoregressive} proposed to used knowledge distillation and a length regulator to predict the end of the generated sign pose sequence, but could not mitigate completely false decoding initiation. The work of Huang~\etal~\cite{Huang2021TowardsFA} leverages a Spatial-Temporal Graph Convolutional Network (ST-GCN)~\cite{yan2018spatial} to explicitly induce a skeletal bias into the decoding step. Although it achieves competitive results, their work requires the use a component trained separately in order to predict the sequences' length.

Although our model is also a non-autoregressive approach, we differ from these last three works by adopting a Generative Latent Optimization (GLO) approach~\cite{bojanowski2018optimizing}. Our strategy is able to provide a well-organized latent space that covers the entire dataset. To decode feature vectors from this rich space, we carefully designed graph pyramid for the facial representation and directly regress the landmarks of facial expressions using a ST-GCN. Moreover, our method pushes forward facial expressions generation for sign language by leveraging sentiment information, which is a key component for correctly understanding the conveyed information. Table~\ref{tab:slp_methods} shows an overview the limitations of some existing methods.

\begin{table}[t!]
\centering
\resizebox{0.99\linewidth}{!}{%
\begin{tabular}{ll}
\toprule
\textbf{Method} & \textbf{Issues/Limitations} \\
\toprule
Stoll~\etal~\cite{stoll2018} & It lacks smooth transitions between signs.\\
Saunders~\etal~\cite{saunders2020progressive}& Error propagation, regression to the mean. \\
Hwang~\etal~\cite{hwang2021non} & It does not consider target sign lengths, false decoding initiation. \\
Huang~\etal~\cite{Huang2021TowardsFA} & It depends on separately trained components. \\
Hwang~\etal~\cite{hwang2022nonautoregressive} & Incomplete mitigation of false decoding initiation. \\
\bottomrule
\end{tabular}
}
\caption{{\bf Limitations}. Overview of the limitations of some existing methods.}
\label{tab:slp_methods}
\end{table}

\vspace{-1em}
\paragraph{Facial Expression Synthesis} 


Virtually all research in human face synthesis lies in the audio-based methods, where input audio is used to synthesize realistic facial human expressions. These methods can be divided into lip-sync~\cite{you_said_that, talking_face, lip_movements, lip_sync_expert}, which focus solely on synchronizing mouth movements with the audio; and full-face expressions~\cite{eskimez, greenwood18, controllable_facial_synth}. 
Cudeiro~\etal~\cite{cudeiro} model speaking style and regress directly FLAME~\cite{FLAME} parameters to 3D animate expressions. Thies~\etal~\cite{thies} take a step further and model separately generalized latent audio expressions and a person-specific 3D model to render a video output. Wu~\etal~\cite{imitating} proposed a model that produces expressive facial results but relies on a style-conditioning video input, which is unsuitable for our task. Moreover, to generate meaningful signs expressions, we need to move the whole face; thus, lip-sync methods do not fit this task. 

\begin{figure*}[t!]
    \centering
    \includegraphics[width=0.95\linewidth]{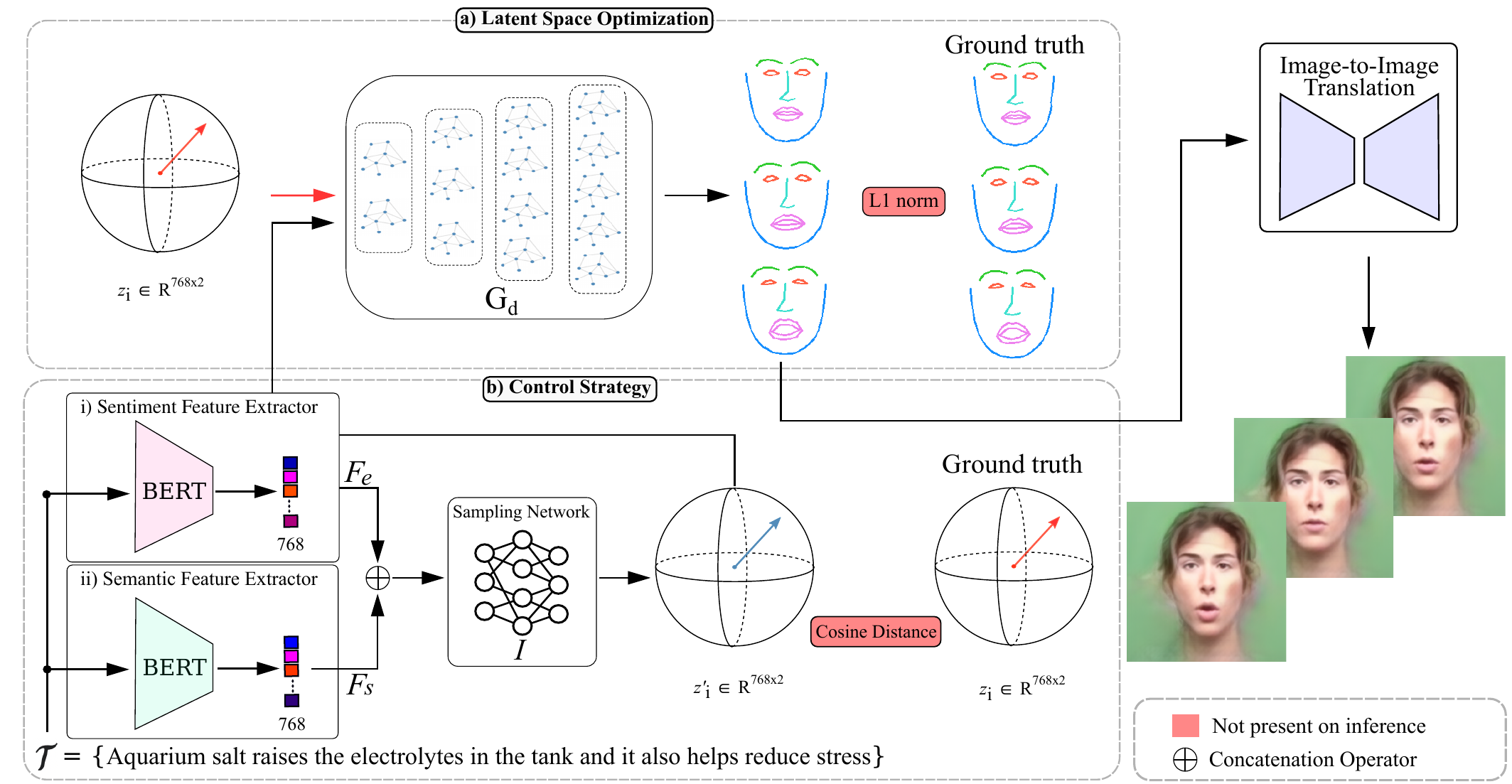}
   \caption{{\bf Overview of our facial expressions synthesis approach}. Our method is composed of two main components: (a) At first, we learn a representation space that contains all the knowledge about the facial expressions in an organized manner; (b) Then we learn how to sample from our space, considering essential aspects in a dialogue, such as emotional state and semantics.}
  
   \label{fig:method}
\end{figure*}

In MakeItTalk~\cite{Yang:2020:MakeItTalk}, a close work to ours, the authors proposed a two-staged deep learning facial expression synthesis model conditioned in audio input and a speaker identifier. In the first stage, the model predicts $68$ facial landmarks displacements by using two separated modules with a combination of LSTM and self-attention mechanisms, modeling short and long-range temporal dependencies. At the second stage, an image-to-image translation method is proposed to generate the final corresponding image output given the predicted facial landmarks and a performer image.

Regarding sentiment awareness, a work that comes close to ours is the EAMM~\cite{eamm}. In this work, the authors propose to transfer the local emotional deformations to an audio-driven talking face with self-learned keypoints and local affine transformations. Their methodology leverages deep learning to map audio representations and extracted poses to unsupervised keypoints and their corresponding first-order dynamics and then extracts emotion-related displacements. The final motions are a linear combination of all motion representations from the two modules.




In this work, we take a step towards modeling human faces to synthesize sign language facial expressions, providing an explicit control mechanism for what is generated, which takes into account the speaker's sentiment. In contrast to the other approaches, our work seeks to explicitly add the bias of elocution of sentiment and semantic encoding that facial expressions have in a specific method for this type of non-manual gesture.

\section{Methodology}

Our method is designed to synthesize a sequence of facial expressions conveying the non-manual component of sign language. Specifically, given a sentence of written text $\mathcal{T}$, we aim to estimate a facial gesture $\mathcal{G}$ composed of a sequence of $N$ facial expressions defined by:
\begin{align}
    \label{eq:expressions}
        \mathcal{G} = \left \{ \mathbf{E_0}, \mathbf{E_1}, \cdots, \mathbf{E_N} \right \} \in \mathbb{R}^{N\times 69 \times 2},
\end{align}
\noindent where ${\mathbf{E_i} = \left \{\mathbf{F_0}, \mathbf{F_1}, \cdots, \mathbf{F_{68}} \right \}}$ is a graph representing the face in frame $i$ and $\mathbf{F_i} \in \mathbb{R}^2$ the 2D image coordinates of the $i$-th node of this graph, \ie, the $i$-th facial landmark. The generated facial gesture is used to feed the Image-to-Image translation module proposed by Yang~\etal~\cite{Yang:2020:MakeItTalk} that generates realistic human facial expressions.

Our approach comprises two main steps, outlined in Figure~\ref{fig:method}. First, our model is trained to learn a meaningful representation space, which compresses information
about the facial expressions in an organized manner,~\ie, a linear interpolation in the space will generate a smooth interpolation of human facial expressions. Second, the model learns to sample from the representation space, considering essential aspects in a dialogue, such as emotional state and semantics. Since we apply a residual Spatio-Temporal Graph Convolutional Network to generate facial gestures from well-organized latent space, our method is simple to train as opposed to other works that use an adversarial training regime. 

\subsection{Latent Space Optimization}

In order to create a well-organized space for human facial expressions, we adopt the approach of Generative Latent Optimization (GLO)~\cite{bojanowski2018optimizing}. Thus, we consider a set of facial gestures $G = \{ \mathcal{G}_0, \mathcal{G}_1, \cdots, \mathcal{G}_n\}$, where $\mathcal{G}_i \in  \mathbb{R}^{N\times 69 \times 2}$. We initialize a set of 
random vectors $\mathcal{Z} = \{ z_0, z_1, \cdots, z_n \} $, where $z_i \in \mathbb{R}^{C \times 2}$. Then, we combine  the dataset of facial gestures with the random vectors, obtaining the dataset $D = \{ (\mathcal{G}_0, z_0), (\mathcal{G}_1, z_1), \cdots,  (\mathcal{G}_n, z_n) \}$. At last, our model jointly learns the $\theta$ in $\Theta$ parameters of a graph convolutional generator $G_{d\theta} : \mathcal{Z} \rightarrow G$ and the optimal random vector $z_i$ for each facial expression by solving:

\begin{equation}
    \min_{\theta \in \Theta} \frac{1}{N} \sum_{i=1}^{n} \left[\min_{z_{i} \in \mathcal{Z}} \ell(G_{d\theta}(z_{i}), \mathcal{G}_i)\right],
\end{equation}

\noindent where $\ell$ is a loss function measuring the reconstruction error from $G_{d\theta}(z_{i})$ to $\mathcal{G}_i$. In our experiments, we adopt the $l_1$ norm; and after each $z$ update in GLO training, we project it onto the unit sphere~\cite{bojanowski2018optimizing}.

\begin{figure*}[t!]
    \centering
    \includegraphics[width=0.95\linewidth]{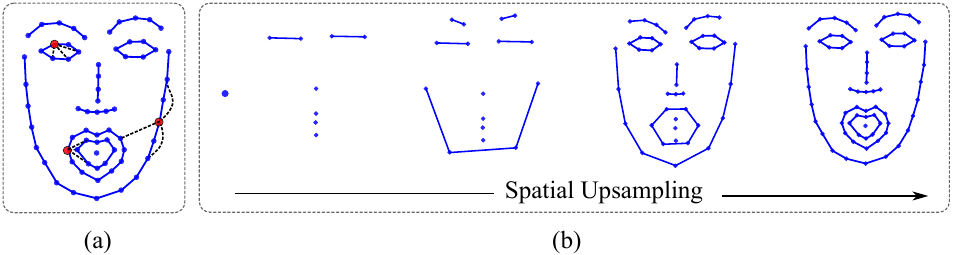}
   \caption{{\bf Face graph topology}. (a) We use the $k$-NN search to determine $k$ connections of each vertex. The figure shows the added edges (black lines) of three vertices (red dots); (b) Spatial upsampling graph pyramid. We designed a sequence of graphs with growing levels of detail. For sake of clarity the $k$-NN edges are omitted.}
   \label{fig:face-graph}
\end{figure*}

\subsubsection{Graph Convolutional Decoding}
\label{subsection:gcd}
Motivated by the representation power of Graph Convolutional Networks (GCNs) in structured data~\cite{yan2018spatial, FERREIRA202111}, our decoder is based on a Residual Spatio-Temporal Graph Convolutional Networks (Res-STGCN). Thus, we use the ST-GCN architecture to explore the local interactions of the face landmarks. Our graph has $|\mathbf{V}| = 69$ vertices and, following Xin~\etal~\cite{xin2021evagcn}, for each vertex $\mathbf{V}$ representing a facial landmark, we use the $k$-Nearest Neighbor ($k$-NN) method to find its $\mathbf{K}$ nearest neighbors. Then, we connect them with edges $\mathbf{Q} = {\{(v_{i},v_{j}) |v_{i},v_{j} \in V\}}$, creating an undirected face graph $\mathbf{W} = (\mathbf{V}, \mathbf{Q})$. Figure~\ref{fig:face-graph}-(a) illustrates the graph topology.

\begin{figure}[!t]
    \centering
    \includegraphics[width=0.98\linewidth]{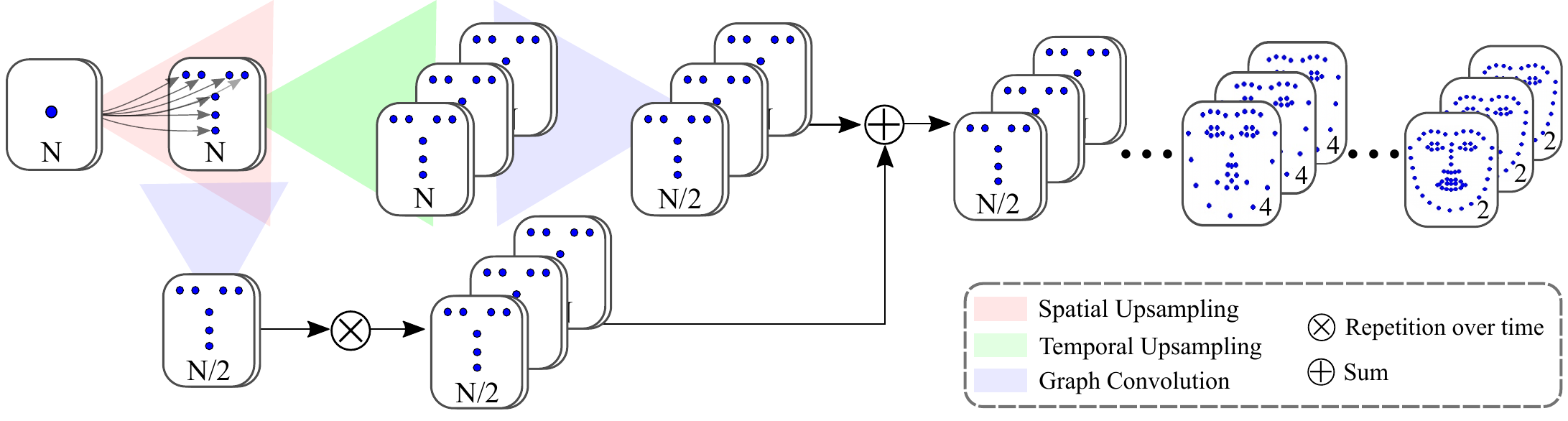}
   \caption{{\bf Graph Convolutional Decoder}. Our decoder network is composed of blocks temporal and spatial upsampling, and graph convolutions layers. The architecture maps from a meaningful representation space to face moving landmarks.}
   \label{fig:decoder-architecture}
\end{figure}

Our decoder network $G_d$ is composed of temporal and spatial upsampling layers, and graph convolutions. In the temporal upsampling layer, we applied transposed 2D convolutions to double the temporal dimension and to enforce the learning of temporal relationships between the graph sequences. We use a residual connection between each temporal upsampling layer. Specifically, the decoder $G_d$ starts with one graph node with $2C$ features from latent space (random values in the beginning). In the subsequent blocks, the upsampling layers compute the features corresponding to face graphs at different scales using $7$, $16$, $43$, and $69$ nodes, respectively. The last layer outputs a graph containing the $(x,y)$ coordinates of each facial landmark. The last decoder layer outputs a sequence of graphs with size $(M, N, V')$, where $M = 2$, $N = 64$, and $V' = 69$. Figure~\ref{fig:decoder-architecture} shows a schematic representation of our architecture.  

For the spatial upsampling, we use the same layer defined by Ferreira~\etal~\cite{FERREIRA202111}. This layer maps a graph $W(V, Q)$ with $V$ vertices and $Q$ edges to a refined graph $W'(V', Q')$ using an aggregation function based on a graph pyramid topology (Figure~\ref{fig:face-graph}-(b)). More precisely, we define $B$ adjacency matrices, where in the $b$-th matrix the edge $(i, j)$ represents a connection between the $i$-th vertex in the new graph $W'$ and the $j$-th vertex in the old graph $W$ if the geodesic distance of $i$ and $j$ in the new graph is equals to $b$. Then, we concatenate these adjacency matrices into the tensor $A^\omega \in \mathbb{R}^{B\times|V'|\times|V|}$ with $b$ starting from $0$. Thus, given the features $\textbf{f}_{j}$ of a node $j$ in graph $W$ and the mapped features $\textbf{f}_{i}$ in a node $i$ in the new graph $W'$, the aggregation function is defined by
\begin{equation}
    \textbf{f}_{i} = \sum_{b, j} A^{w}_{bij}\textbf{f}_{j}.
\end{equation}
Each edge in $A^\omega$ has trainable weights that enable the network to optimize the feature vectors from the upsampling operations. Additionally, a graph convolutional layer~\cite{yan2018spatial} is included after each type of upsampling layer in order to learn the spatio-temporal relationship between the graphs.

\subsection{Control Strategy}
\label{subsection:controlstrategy}

To sample from a GLO model, we have to fit a single full-covariance Gaussian to the $z$ and apply a Gaussian~\cite{bojanowski2018optimizing}. As a drawback, this process does not enable explicit control of expression is generated. To overcome this limitation, we adopt a Sampling Network $I$ that provides control over the GLO model by sampling a latent vector with the desired characteristic.


Moreover, we explore the correlation between semantics and the sentiment of a sentence with facial expressions, by training the neural network $I$ with both semantic and semantic features as inputs. In other words, let $\mathcal{T}$ be a written text sentence corresponding to an optimized random vector $z$. We extract semantic ($F_s$) and sentiment ($F_e$) features and train a network to approximate a representation for $z$.
\vspace{-0.1cm}
\paragraph{Sampling Network $I$}
Our Sampling Network $I$ receives as input two sets of features: semantic ($F_s$) and sentiment ($F_e$) features. Given a sentence $\mathcal{T}$, the Sentence Transformers model~\cite{reimers-2019-sentence-bert} extracts the representations of the sentence,~\ie, a semantic feature $F_{s} \in \mathbb{R}^{768}$. Our strategy to extract sentiment features is inspired by the work of Yin~\etal~\cite{yin2020sentibert}. We use the SentiBert~\cite{yin2020sentibert} feature extractor to train a classifier on the SemEval2018 Dataset~\cite{SemEval2018Task1} aiming to learn the correlation of a sentence and its emotion label. After training the emotion classifier, we decoupled the encoder from the architecture to use it as a feature extractor. This encoder outputs a sentiment feature vector $F_{e} \in \mathbb{R}^{768}$ associated with the input sentence.

\begin{figure}[t!]
    \includegraphics[width=0.85\linewidth, center]{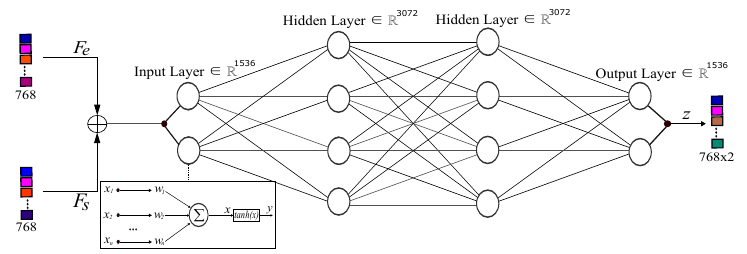}
   \caption{{\bf Sampling Network}. Our sampling network is composed of fully connected layers. The architecture maps between the concatenated sentiment $F_e$ and semantic features $F_s$ to the learned representation space $z$.}
   \label{fig:nn}
\end{figure}

Since the input and output are sentences and there is no spatial correlation, our architecture comprises four fully connected linear layers, each followed by a hyperbolic tangent function as represented in Figure~\ref{fig:nn}. In the training, our decoder $G_d$ receives a normalized vector as input since we project the $z$'s into the unit sphere. Besides, the output is a high-dimensional vector. Therefore, the Sampling Network $I$ is trained by computing the Cosine Distance $L_c$ between the optimized random vector $z_i$ (ground truth) and the regressed $z_i' = I(F_s, F_e)$ as

\begin{equation}
\label{eq:cosine}
L_c(I(F_s, F_e), z_i) = \left[ 1 - \frac {I(F_s, F_e) \cdot z_i}{||I(F_s, F_e)|| \cdot ||z_i||} \right].
\end{equation}


Finally, the inference comprises four steps. First, given a sentence of text $\mathcal{T}$, we extract semantic ($F_s$) and sentiment ($F_e$) features using the same process described for feature extraction in training our Sampling Network $I$ within the Control Strategy component. Second, we feed our Sampling Network $I$ (\ref{subsection:controlstrategy}) with the extracted features to produce the estimated latent vector. Third, we project the estimated latent vector into the unit sphere. At last, we feed our decoder $G_d$ (\ref{subsection:gcd}) with the normalized latent vector to produce human facial expressions.

\begin{table}[t!]\centering

\begin{tabular}{cccccc}\toprule

    \multicolumn{3}{c}{Phoenix14T} & \multicolumn{3}{c}{How2Sign} \\
    \cmidrule(lr){1-3}\cmidrule(lr){4-6}
    Speaker &\multicolumn{1}{c}{Train} &\multicolumn{1}{c}{Test}& Speaker &\multicolumn{1}{c}{Train}&\multicolumn{1}{c}{Test} \\
    \cmidrule(lr){1-1}\cmidrule(lr){2-2}\cmidrule(lr){3-3}\cmidrule(lr){4-4}\cmidrule(lr){5-5}\cmidrule(lr){6-6}
    

    $1$ &$1{.}234$ & $138$ & $5$ &$7{.}519$ & $200$\\
    $5$ &$940$ & $110$ & $8$ & $6{.}083$ & $192$\\
    $4$ &$921$ &$66$ & $3$ & $1{.}119$ & $336$\\
    $8$ &$723$ &$50$ & $1$ & $568$ & $123$\\
    
    \bottomrule
\end{tabular}
\caption{\textbf{Datasets splits.} The top four signers based on data volume for each dataset (Phoenix14T on the left, How2Sign on the right). All data was split into train and test sets.}\label{tab:datasets_splits}
\end{table}
\section{Experiments}


\paragraph{Dataset and preprocessing}

We assess the effectiveness of our method in two publicly available datasets in the field of Sign Language Production, the How2Sign~\cite{Duarte_CVPR2021}, a multimodal and multiview continuous American Sign Language (ASL) dataset and the PHOENIX14T~\cite{phoenix}, a common benchmark in German Sign Language (GSL). Since facial expressions are extremely personal and might vary drastically from person to person, we built a person-specific training and test set, choosing the four speakers with more samples available from each dataset as shown in Table~\ref{tab:datasets_splits}. In sequence, we sample the ground truth frames uniformly according to the video length keeping all sequences the same size. Following Vonikakis~\etal~\cite{frontalization}, we frontalize the 2D facial landmarks keeping the person's identity. Additionally, we applied a low-pass filter~\cite{oneeurofilter} to smooth out the jittering caused by the OpenPose~\cite{openpose} estimator.



\paragraph{Parameters Setting and processing time}
The channel size of the random vectors, $F_e$ and $F_s$ are defined as $C = 768$. The spatial dimension of the initial level of the graph pyramid is set to $V = 1$, the number of neighbors to generate the face graph is $K = 3$ and the facial expressions output sequence size chosen as $N = 64$. For the maximum geodesic distance in the upsampling layers, we defined $B = 2$. We trained our model for approximately one day on setup with an Intel Xeon CPU E5-2620, an NVIDIA Titan RTX GPU 24GB VRAM, and 86.5GB RAM. The inference time of a unitary batch is $0.02$ seconds.

\paragraph{Face animations}
In order to produce real animated faces, we use the image-to-image translation module proposed by Yang~\etal~\cite{Yang:2020:MakeItTalk}. Given an input identity image $\mathcal{I}$ and the predicted facial expressions $\mathcal{G'}$, we generate a sequence of images $\mathcal{I'} = \left \{ \mathcal{I'}_0, \mathcal{I'}_1, \cdots, \mathcal{I'}_n \right \}$ representing the facial animation, and $n$ being the number of frames. It is worth mentioning that any other method capable of transferring motion from facial landmarks can be applied. 

\begin{table}[t!]
    \centering
    \resizebox{\linewidth}{!}{%
        \setlength{\tabcolsep}{0.25cm}
        \begin{tabular}{@{}clccccccc@{}}
            \toprule
            \multirow{3}{*}{\bf Metrics $(\downarrow)$}
            &  \multicolumn{1}{c}{}
            &  \multicolumn{7}{c}{{\bf Methods}}\\ \cmidrule{3-9}
            &  \multicolumn{1}{c}{} 
            &  {\centering \textit{Ours}}
            &  {\centering Wo/Sent}
            &  {\centering Wo/Sem}
            &  {\centering Wo/SN} 
            &  {\centering Wo/GLO} 
            &  {\centering Wo/GCN}
            &  {\centering Wo/k-NN} \\ \midrule
            
            & \hspace*{-2.2cm} \textit{Fréchet Expression Distance} & \textbf{6.10} & 7.91 & 7.25 & 7.88 & 21.47 & 45.41 & 9.88 \\

            & \hspace*{-2.2cm} \textit{Fréchet Video Distance} & \textbf{120.90} & 121.20 & 121.42 & 127.97 & 231.25 & 288.91 & 148.20 \\
            \bottomrule
        \end{tabular}
    }
    \caption{{\bf Ablation study}. Our complete model (\textit{Ours}) achieves overall better results compared with other alternatives (best in bold).}
    \label{table:ablation_result}
    \vspace{-0.5cm}
\end{table}

			


            
            
            

\subsection{Evaluation metrics}

We evaluate the quality of one synthesized sequence of facial expressions in terms of a set of quantitative metrics. To the best of our knowledge, there was no model for extracting features from facial expressions for sign language to apply the FID~\cite{FID} metric. Therefore, following~\cite{Yoon2020Speech}, we trained an autoencoder on the How2Sign~\cite{Duarte_CVPR2021} and PHOENIX14T~\cite{phoenix} datasets and used the encoder part as a feature extractor to measure the Fréchet Expression Distance (FED) between generated results and the ground truth. We also use the Fréchet Video Distance (FVD) \cite{unterthiner2018towards} to measure the quality of our animations when compared to ground truth videos. Moreover, to assess the quality of specific regions of the face important for sign language, following Yang~\etal\cite{Yang:2020:MakeItTalk}, we computed the Euclidean distance between the mouth, eyebrows, and jaw-lips between the sequence produced and the ground truth.

\subsection{Ablation Study}

In order to verify the contributions of various components of the model on the overall facial expression generation, we design six baselines using only Speaker $8$ data from the How2Sign dataset. Speaker 8 was selected because its number of samples is closest to the average across all speakers. The six baselines showing the effect of adding or removing specific components are as follows:

i) \textbf{Wo/Sem}: The sampling network without the semantic feature; ii) \textbf{Wo/Sent}: The sampling network without the sentiment feature; iii) \textbf{Wo/SN}: Remove the sampling from the optimized latent space with a neural network, we apply a heuristic based on the two closest sentences in semantic and sentiment and interpolate between them to generate new samples; iv) \textbf{Wo/GLO}: Remove the GLO model and train the framework end-to-end; v) \textbf{Wo/GCN}: Remove the GCN-based decoder and train the framework with a standard MLP; vi) \textbf{Wo/k-NN}: Remove the k-NN method to generate connections in face graph topology.



Table~\ref{table:ablation_result} shows that the model performance drops in all metrics without the semantic information (Wo/Sem). This result indicates that adding information related to the structure and the semantic of the words in a sentence improves the learning of a correlation between the written text and sign language facial expressions.


\begin{table*}[tb]
	\centering
	\resizebox{0.99\linewidth}{!}{%
    \setlength{\tabcolsep}{0.2cm}

		\begin{tabular}{llrcccc}
		    \toprule
			\multirow{3}{*}{\bf Metrics $(\downarrow)$}
			&  \multicolumn{1}{c}{}
			&  \multicolumn{5}{c}{{\bf Methods}}\\ \cmidrule{3-7}
			&  \multicolumn{1}{c}{} 
			&  {\centering {\it Ours}}
			&  {\centering Progressive Transformers~\cite{saunders2020progressive}}
                &  {\centering NSLP-G~\cite{hwang2021non}}			
                &  {\centering MakeItTalk~\cite{Yang:2020:MakeItTalk}} 
                &  {\centering EAMM~\cite{eamm}} 
                    \\ \midrule
			
			& \hspace*{-2.2cm} Avg. Landmarks Dist. & \textbf{0.1903} & 0.9410 & 0.2988 & 0.3332 & 0.3089 \\

			& \hspace*{-2.2cm} Avg. Jaw-Lips Dist. & \textbf{0.1730} & 0.7809 & 0.2846 & 0.3101 & 0.2967\\

			& \hspace*{-2.2cm} Avg. Eyebrows Dist. & \textbf{0.0573} & 0.3101 & 0.0771 & 0.0728 & 0.0718\\
            
            & \hspace*{-2.2cm} Avg. Mouth Dist. & \textbf{0.0927} & 0.5021 & 0.1541 & 0.2005 & 0.1754\\
            \midrule
            
            & \hspace*{-2.2cm} \textit{Fréchet Expression Distance (FED)} & \textbf{14.49} & 29.04 & 28.02 & 149.11 & 132.37\\
            & \hspace*{-2.2cm} \textit{Fréchet Video Distance (FVD)} & \textbf{136.41} & 237.11 & 245.74  & 301.22 & 286.66\\
        	\bottomrule        
\end{tabular}
	}
     \caption{{\bf Quantitative evaluation in the PHOENIX14T Dataset}. Comparison with our baselines according the average distances of jaw-lips, mouth, eyebrow, overall landmarks, Fréchet Expression Distance and Fréchet Video Distance metrics (best in bold).}
    	\label{table:baselines_phoenix}
\end{table*}

\begin{table*}[tb!]
	\centering
	\resizebox{0.99\linewidth}{!}{%
    \setlength{\tabcolsep}{0.2cm}

		\begin{tabular}{llrcccc}
		    \toprule
			\multirow{3}{*}{\bf Metrics $(\downarrow)$}
			&  \multicolumn{1}{c}{}
			&  \multicolumn{5}{c}{{\bf Methods}}\\ \cmidrule{3-7}
			&  \multicolumn{1}{c}{} 
			&  {\centering {\it Ours}}
			&  {\centering Progressive Transformers~\cite{saunders2020progressive}}
                &  {\centering NSLP-G~\cite{hwang2021non}}			
                &  {\centering MakeItTalk~\cite{Yang:2020:MakeItTalk}} 
                &  {\centering EAMM~\cite{eamm}} 
                    \\ \midrule
			
			& \hspace*{-2.2cm} Avg. Landmarks Dist. & \textbf{0.1040} & 0.7540 & 0.2075 & 0.2444 & 0.2211\\

			& \hspace*{-2.2cm} Avg. Jaw-Lips Dist. & \textbf{0.0924} & 0.6486 & 0.1995 & 0.2388 & 0.1998\\

			& \hspace*{-2.2cm} Avg. Eyebrows Dist. & \textbf{0.0351} & 0.2062 & 0.0497 & 0.0489 & 0.0491\\
            
            & \hspace*{-2.2cm} Avg. Mouth Dist. & \textbf{0.0679} & 0.4544 & 0.1106 & 0.1768 & 0.1463\\
            \midrule
            
            & \hspace*{-2.2cm} \textit{Fréchet Expression Distance (FED)} & \textbf{9.52} & 23.05 & 21.70 & 129.19 & 111.24\\
            & \hspace*{-2.2cm} \textit{Fréchet Video Distance (FVD)} & \textbf{115.99} & 227.35 & 217.12  & 289.92 & 254.91\\
        	\bottomrule        
\end{tabular}
	}
     \caption{{\bf Quantitative evaluation in the How2Sign Dataset}. Comparison with our baselines according the average distances of jaw-lips, mouth, eyebrow, overall landmarks, Fréchet Expression Distance and Fréchet Video Distance metrics (best in bold).}
    	\label{table:baselines_result}
	\vspace{-0.5cm}
\end{table*}



			


            
            
            

By removing the sentiment features (Wo/Sent), the model performance also degraded in all the metrics. This result shows that the sampling network strategy of encoding the emotional content of the text to sample from a meaningful space of facial expressions leads to more consistent overall results. Finally, aiming to test the effectiveness of our sampling network, we design a heuristic (Wo/SN). This heuristic uses the two closest sentences to the test instance in the sentiment and semantic space by the cosine distance and interpolates between them to find a correspondent representation. According to our results, using a representation learned by a model generates better results than using a heuristic based on a cosine distance. From this result we draw the observations that our network can learn an effective mapping between the semantic and sentiment features and the optimized representations from our representation space. According to the experiments, removing the GLO framework (Wo/GLO) leads to a drop in performance. This can be explained by the fact that GLO creates a more meaningful latent space~\cite{bojanowski2018optimizing} when compared to parametric (end-to-end) approaches. Similarly, removing the GCN-based decoder (Wo/GLO) also leads to a performance drop. This is due to the GCN capacity to leverage geometrical properties embedded in face topology graph, making the method better correlate the expressions with the text than a simple standard MLP. At last, we found that removing k-NN (Wo/k-NN) to create additional edges between different parts of the face yields worse results once this topology is designed to perform spatial operations.


\subsection{Quantitative Evaluation}


We selected four state-of-the-art methods to evaluate the performance of our method both quantitatively and qualitatively, namely, Progressive Transformers~\cite{saunders2020progressive}, NSLP-G~\cite{hwang2021non}, MakeItTalk~\cite{Yang:2020:MakeItTalk}, and EAMM~\cite{eamm}. Progressive Transformers and NSLP-G are both strong baselines in the SLP area, and their codes are publicly available. For the sake of a fair comparison, since these baselines were not originally trained to generate only facial expressions, we trained them on the same setup used in our method. MakeItTalk is a facial synthesis method capable of learning the entire facial expression considering the correlation between all facial elements, which brings it close to our work. EAMM is a talking-head method that learns to generate facial movements from audio while incorporating emotional bias into the generation process, which makes it similar to our work. Since these talking-head methods only requires an identity image for generating a output, we do not finetune it on person-specific data. Although MakeItTalk and EAMM are not sign language production methods, we added them as baselines to demonstrate that, as far as facial expressions for sign language are concerned, it is not enough to convert text to audio and audio to facial expressions because these non-manual gestures have very different purposes in sign language communication, such as applying punctuation and emphasizing signs, and uses different parts of the face such as the eyebrows and cheeks to perform this function.


Table~\ref{table:baselines_phoenix} shows the average results for Speakers $1$, $4$, $5$ and $8$ in the PHOENIX14T Dataset~\cite{phoenix}. Similarly, Table~\ref{table:baselines_result} presents the average results for Speakers $1$, $3$, $5$ and $8$ in the How2Sign Dataset. We can see that our approach achieves superior performance in all metrics, particularly in the FED metric. These results indicate that our method is capable of generating non-manual gestures that are part of a distribution closer to the real expressions once it uses a carefully designed architecture that considers crucial aspects of facial expressions' role in sign language and leverages geometrical properties of facial gestures. Moreover, these results demonstrate our method's ability to work effectively with different speakers and handle significant variations in the quality of the estimated landmarks used during our training process.

\begin{figure}[!t]
    \centering
 	\includegraphics[width=1.\linewidth]{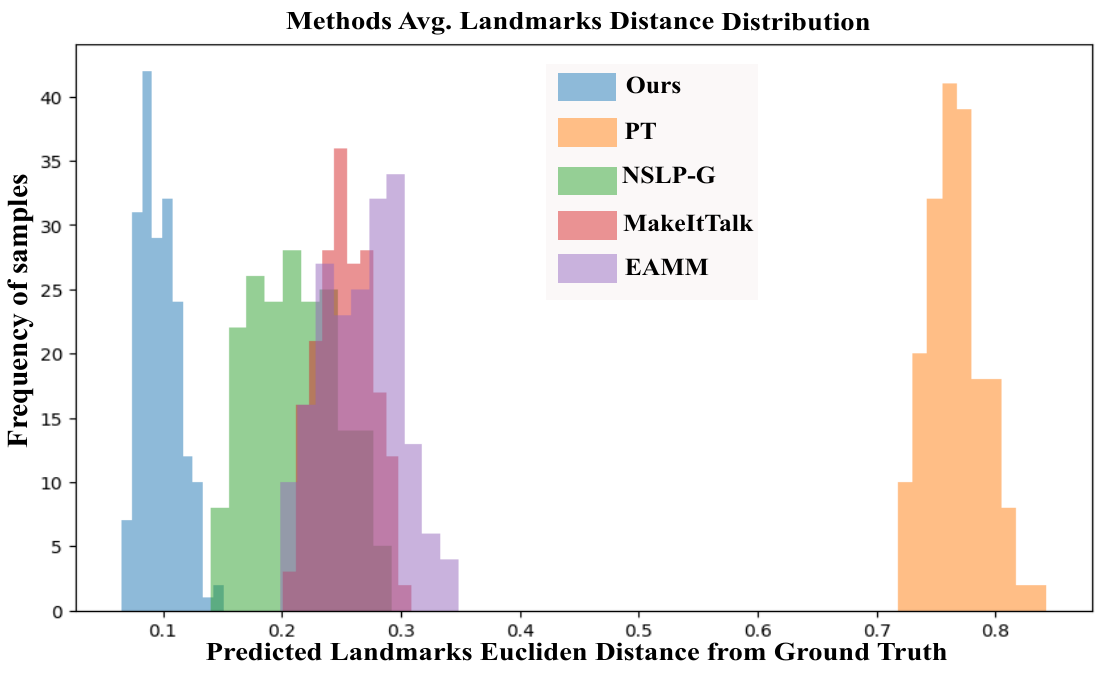}
 	\caption{\textbf{Avg. Landmarks Distance Distribution from Ground Truth} For each method, we show the Avg. Landmarks Distance prediction distribution.}
 	\label{fig:distplot}
\end{figure}

In Figure~\ref{fig:distplot}, for each method, we report the Avg. Landmarks Distance distribution using How2Sign Speaker 8. As we can see, the distance for samples generated by our method are closer to ground truth when compared to other approaches. This result suggests that using a non-autoregressive architecture based on graphs empower our model to generate better facial gestures when compared to a Transformers based method~\cite{saunders2020progressive}. Moreover, our approach also outperformed the NSLP-G~\cite{hwang2021non}, indicating that using the GLO framework to generate a meaningful representation space for facial expressions instead of a Variational Autoencoder generates results closer to the real expressions made by sign language speakers. From the results, it is noteworthy that when generating facial expressions for sign language it is not enough to only convert text to audio and expressions using a Facial Expression Synthesis method such as MakeItTalk~\cite{Yang:2020:MakeItTalk} and EAMM~\cite{eamm} since facial gestures in sign language have the function of expressing emotions while encoding semantics using different parts of the face.


\begin{figure*}[!t]
    \centering
 	\includegraphics[width=0.9\linewidth]{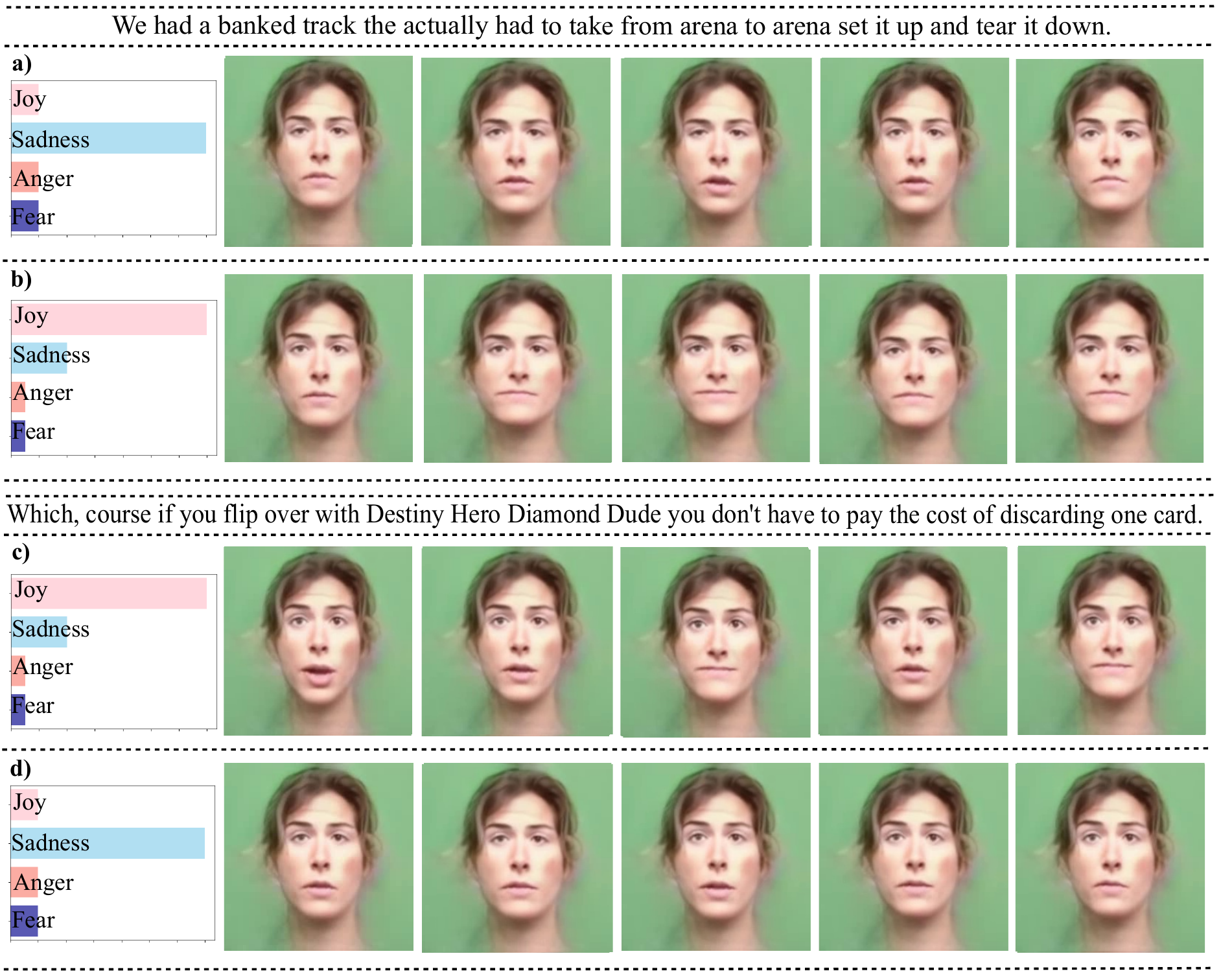}
 	\caption{\textbf{Facial expression synthesis for sign language.} Given a sentence of written text, our method generates distinct non-manual gestures (a, b, c and d) according to the sentiment information used as input to the model. All expressions in the image were automatically generated by our method.}
 	\label{fig:expressions_sentiment_sp82}
\end{figure*}

\subsection{Qualitative Evaluation}

Aside from the quantitative assessment, we also evaluate our approach qualitatively since the visual inspection of the decoded facial expressions also concurs with the quantitative analysis. We study the visual relationship between the emotional content of a text and the correspondent-generated facial expressions.

Figure~\ref{fig:expressions_sentiment_sp8} shows that, for the same sentence, our method can  generate different non-manual gestures according to the given sentiment information. In Figure~\ref{fig:expressions_sentiment_sp8}-(a) and Figure~\ref{fig:expressions_sentiment_sp8}-(d), we show that when feeding the model with a representation of sentiment labeled as \textit{joy}, expressions with a happier countenance are generated. On the other hand, when a representation classified as \textit{anger} is used, as shown in Figure~\ref{fig:expressions_sentiment_sp8}-(b) and Figure~\ref{fig:expressions_sentiment_sp8}-(c), the method generates faces with expressions containing a frowned eyebrow and a more opened mouth, resembling a negative expression. Similarly, in Figure~\ref{fig:expressions_sentiment_sp82}-(a) and Figure~\ref{fig:expressions_sentiment_sp82}-(d), we show that when the model is fed with an embedding related to \textit{sadness}, expressions with dropped eyelids and lip corners pulled down are generated, resembling a sad face. On the other hand, in Figure~\ref{fig:expressions_sentiment_sp82}-(b) and Figure~~\ref{fig:expressions_sentiment_sp82}-(c), when using an embedding labeled as \textit{joy}, the method produces non-manual gestures with lip corners raised and a more opened eye, showing a happy countenance.

\begin{figure*}[t!]
    \centering
 	 \includegraphics[width=0.85\linewidth]{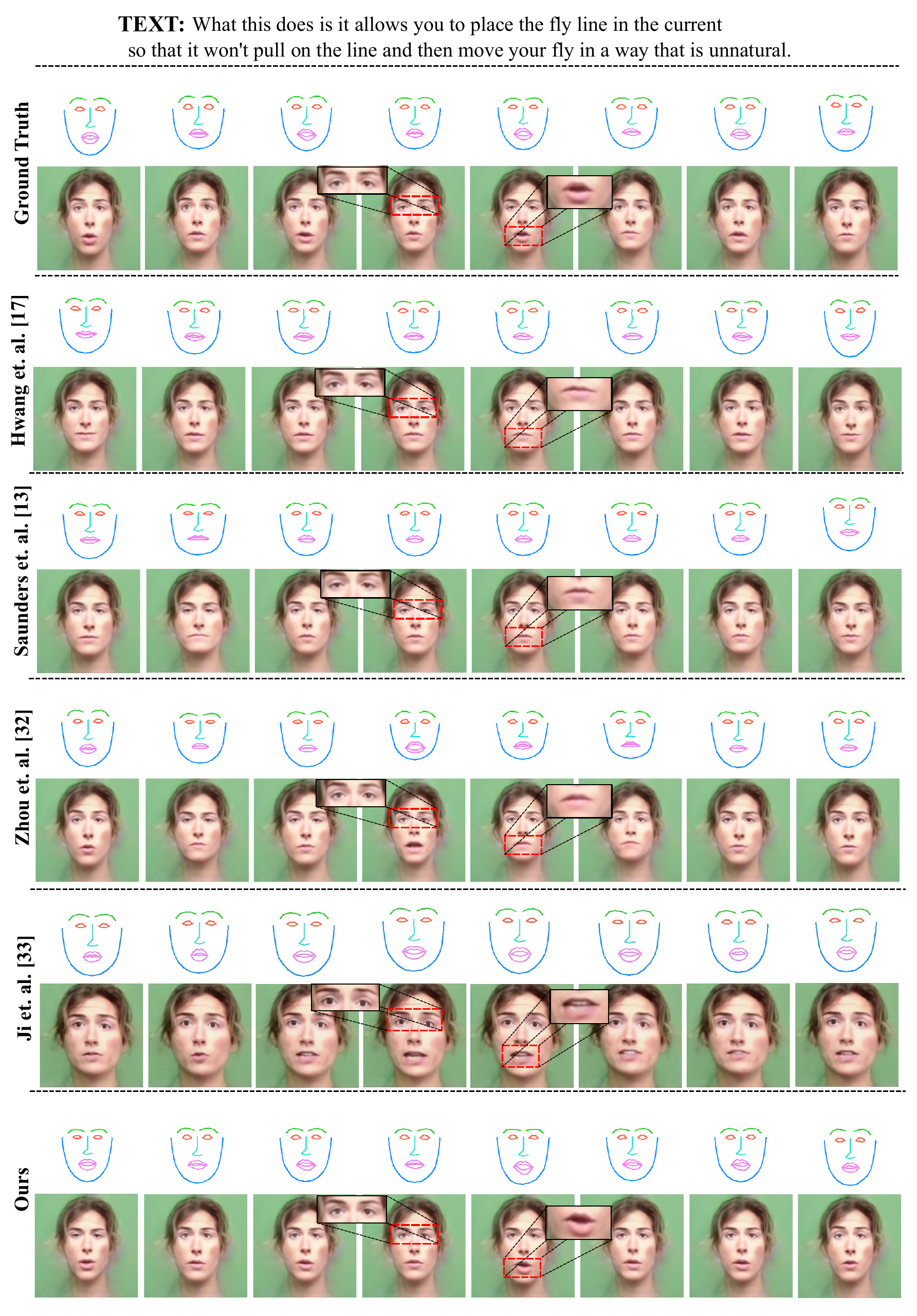}
 	\caption{\textbf{Qualitative comparison.} The first row shows the input text and the rows below present the results for the competitors (landmarks and generated images). The red squares highlight important regions for sign language and show the generation quality.}
 	\label{fig:baseline_expressions_sp8}
 \end{figure*}
 

\begin{figure*}[ht!]
    \centering
 	\includegraphics[width=0.79\linewidth]{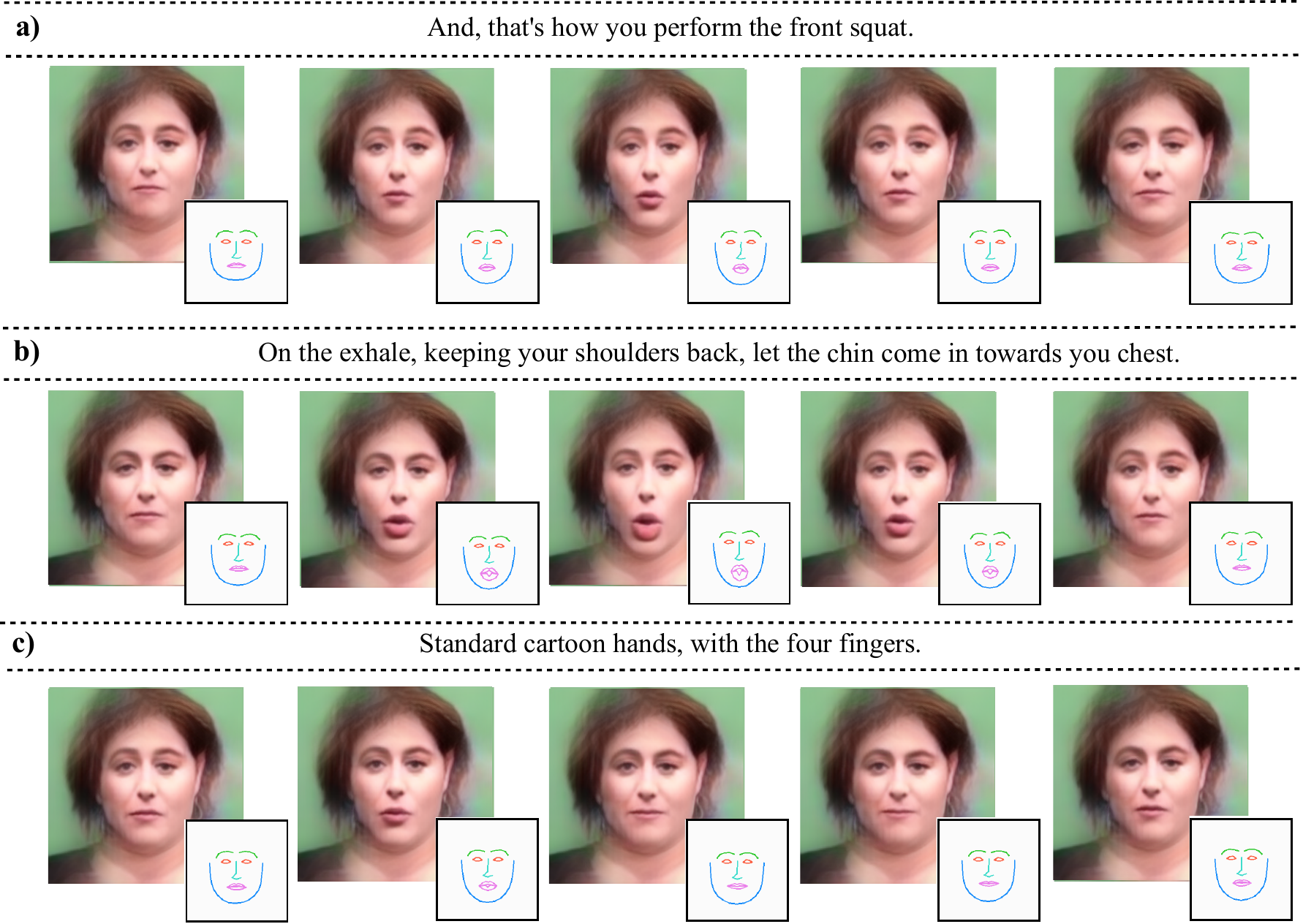}
 	\caption{\textbf{Qualitative Results.} Expressions generated by our method for Speaker 5. For each item (a, b, and c), we have the input text and our results (landmarks and generated images).}
 	\label{fig:expressions_sp5}
\end{figure*}

\begin{figure*}[ht!]
    \centering
 	\includegraphics[width=0.79\linewidth]{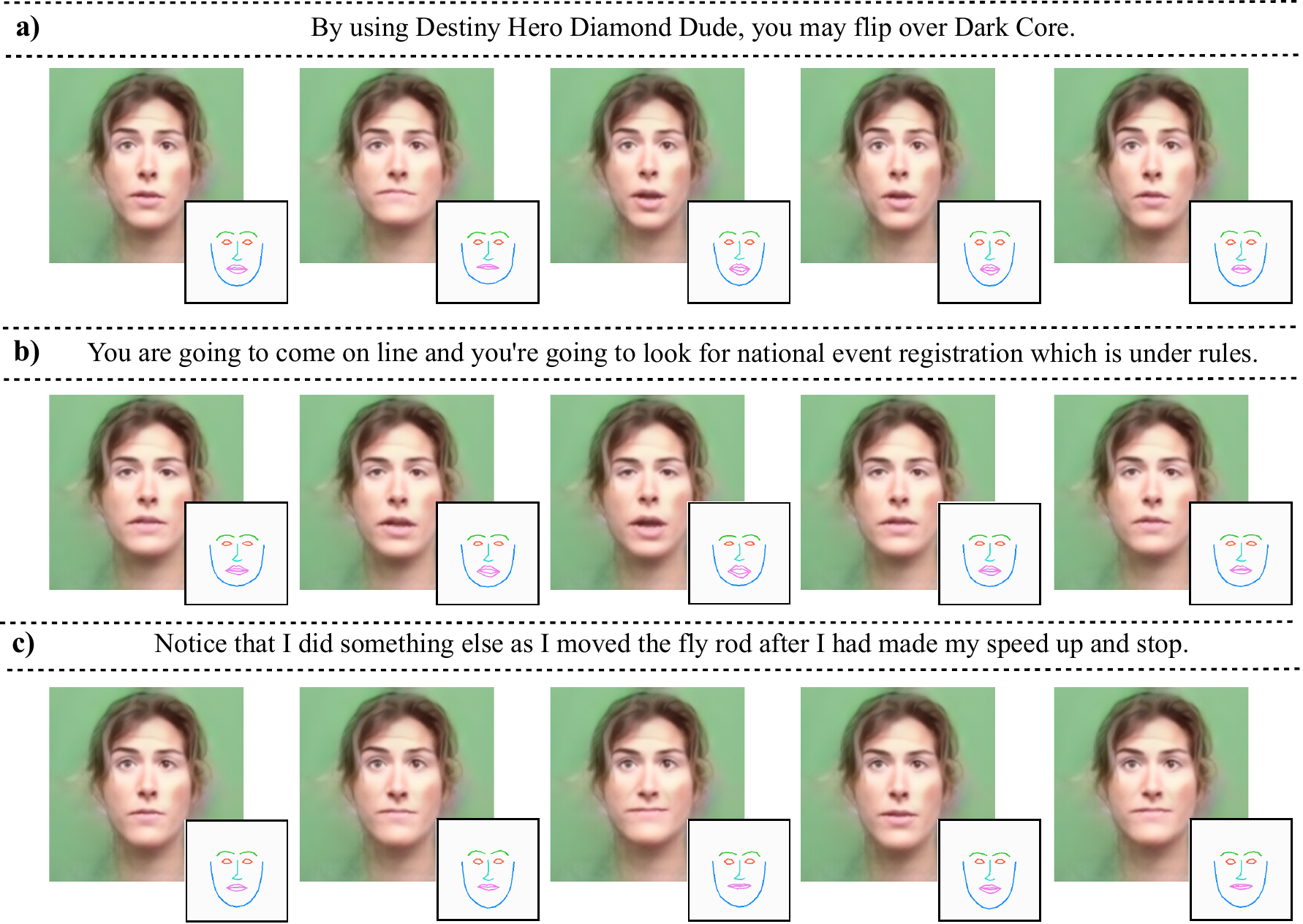}
 	\caption{\textbf{Qualitative Results.} Expressions generated by our method for Speaker 8. For each row (a, b, and c), we show the input text and our results (landmarks and generated images).}
 	\label{fig:expressions_sp8}
\end{figure*}

Furthermore, we perform a visual comparison with the work of Saunders~\etal~\cite{saunders2020progressive}, Yang~\etal~\cite{Yang:2020:MakeItTalk}, Hwang~\etal~\cite{hwang2021non}, and Ji~\etal~\cite{eamm}. Figure~\ref{fig:baseline_expressions_sp8} shows a sample of frames generated from the test set. Although all methods were capable of generating movement, only ours provided the motion in similar temporal occurrence to ground truth (highlighted face regions in Figure~\ref{fig:baseline_expressions_sp8}). This result indicates that our network learned the right moment to make the facial gestures given a sentence, indicating it learned to model the role of facial expressions. Figures \ref{fig:expressions_sp5} and \ref{fig:expressions_sp8} shows that our method generates facial expressions with different levels of oral gestures and eyebrow movements. In Figures \ref{fig:expressions_sp5}-(a) and \ref{fig:expressions_sp8}-(a), for instance, the expressions are composed of a more raised eyebrow and a mouth with less opening, while in Figures \ref{fig:expressions_sp5}-(b) and \ref{fig:expressions_sp8}-(b) the faces are frowning with a more opened mouth. The approach can also produce more subtle and finer movements such as in Figures \ref{fig:expressions_sp5}-(c) and \ref{fig:expressions_sp8}-(c), where a softer oral gesture is produced together with raised eyebrows, indicating that the method generates facial gestures of different aspects.

\begin{figure*}[!t]
    \centering
 	\includegraphics[width=0.9\linewidth]{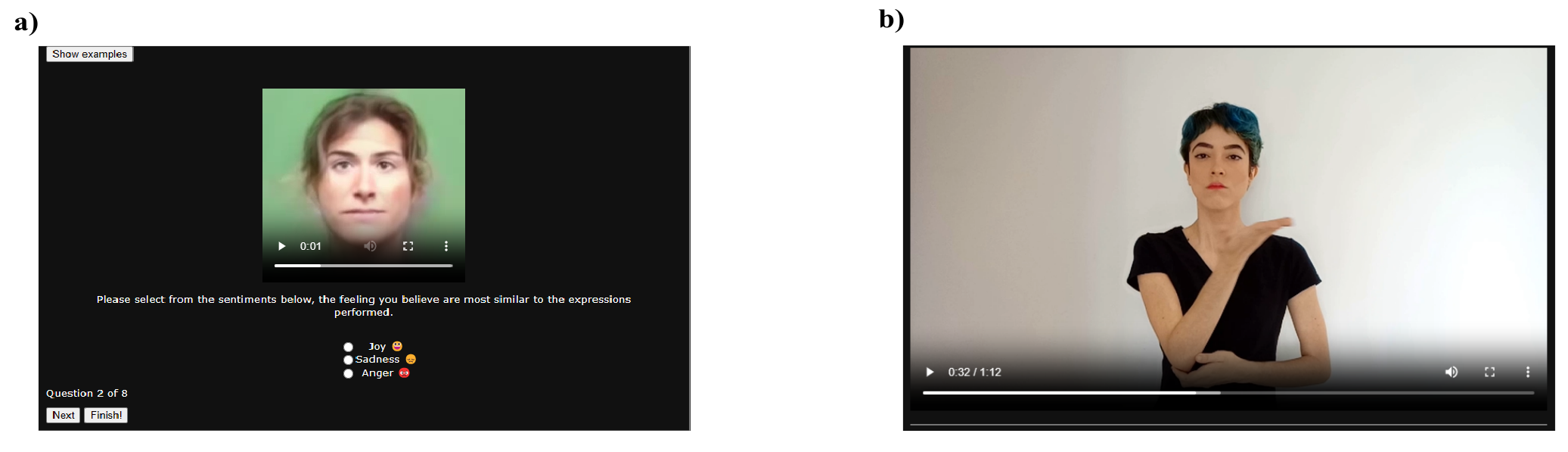}
 	\caption{\textbf{User study interface.} We developed a web interface in order to conduct our user study. The figure presented in (a), shows an example of the options presented to the user for a given video. In (b), we show an example of the accessibility videos of the website to assist sign language speakers carry out the study.}
 	\label{fig:interface}
 \end{figure*}

 \subsection{User study}

We also carried out a user study to verify our method's ability to generate facial expressions that are faithful to the sentiment given as input. Examples of the web interface developed for the study are shown in Figure~\ref{fig:interface}. In this study, we gathered $34$ users of which $5$ were sign language speakers. From these, $41\%$ declared male, $53\%$ female, and $6\%$ did not declare. Our participants had an average age of $34$ years. For oral language speakers, we also collected information regarding their experience with sign languages. $62\%$ had no experience with sign languages, $27\%$ had basic knowledge, $3\%$ had intermediate experience and $8\%$ did not declare. All participants watched four-second randomly ordered videos containing facial animations generated automatically by our method. They were asked to select one option from the list of feelings (\textit{joy}, \textit{sadness}, \textit{anger}) that they felt most resembled the expressions performed in the videos. The list of feelings comprehends all the emotions contained on our test dataset. There were four types of videos generated by our method in the study:

\begin{enumerate}
    \item Videos with the original sentiment label: The purpose of these videos is to assess whether our method is capable of reproducing human-recognizable emotions that match the original sentiment of the sentence, without any kind of exchange;
    \item Videos with a different sentiment label than the original (\textit{e.g.}, a video with an original joy label swapped to sadness): These videos aim to verify whether, in addition to being able to reproduce expressions with emotions recognized by humans, our method is sensitive to the input of sentiment to the point of being able to reproduce an emotion distinct from the original sentiment of the sentence;
    \item Videos with a different sentiment embedding, but with the same sentiment class (\textit{e.g.}, a video with the original joy label swapped to another joy label): Similar to the second type of video, it also seeks to verify the sensitivity to the sentiment of the method, but the objective is to evaluate if, for different embeddings of the same class, the emotions of the generated expressions are consistent with each other;
    \item Video generated using landmarks from the ground truth annotation (Placebo): The purpose of the placebo is to filter inattentive users and add more robustness to the test. Since these are videos generated from groundtruth,  users should hit most of the videos of this type since they are easier when compared to the others. We provided placebos for the \textit{joy} and \textit{anger} emotions.
\end{enumerate}

\begin{figure}[!t]
    \centering
 	\includegraphics[width=1.\linewidth]{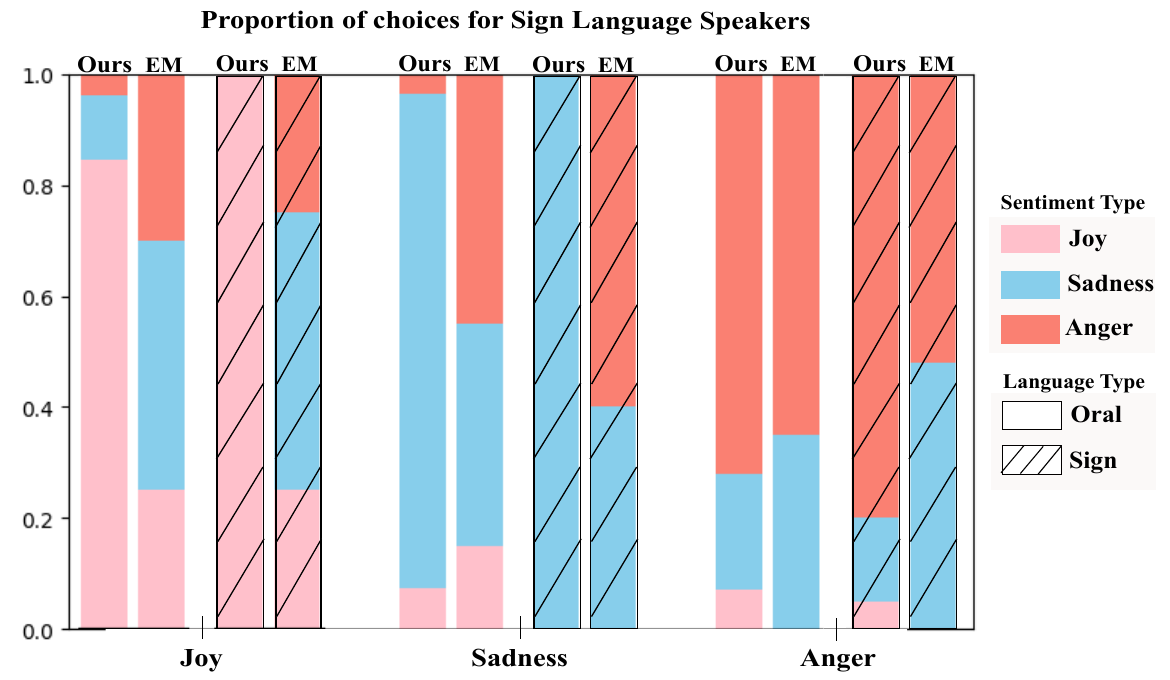}
 	\caption{\textbf{User study results.} For each method, we show the proportion of sign and oral language users' choices for each type of sentiment in the study.}
 	\label{fig:user_study}
 \end{figure}

In addition, to compare our method's sentiment-awareness, we include examples from the work of Ji~\etal~\cite{eamm} since the proposed method is also sentiment-aware. We showed three videos generated by our method, three videos generated by the EAMM~\cite{eamm} and two groundtruth videos. To analyze the survey responses, we calculated the average number of choices for each type of sentiment present in the study. In Figure~\ref{fig:user_study}, we show the generated samples results for both methods for sign and oral language speakers. We can see that our method achieved better results in recognizing the generated emotions in both languages. Particularly for the sign language speakers, we can see that our method obtained better results when compared to the oral language result, while for the EAMM method, we can observe the opposite trend. As for the placebo samples, $89.4\%$ of the users classified \textit{joy} correctly and $85.8\%$ for \textit{anger}. We also test our results to reject the null hypothesis $H_{o}$ that there is no difference between the proportion of choices for each type of emotion label on our videos. Using the Chi-Squared Test ~\cite{chisquared}, considering a significance level of $0.05$, we reject the null hypothesis with $p_{joy}=1.04x10^{-9}$, $p_{sadness}=9.67x10^{-11}$, $p_{anger}=2.24x10^{-5}$.

\section{Limitations} 

Although our method holds state-of-the-art results, it has a few limitations. Since our goal is to explore the use of sentiment representations in generating facial gestures, we do not generate the manual components. Integrating these components and synchronizing it would be the objective of future work. Furthermore, our method uses 2D data only. This limits our model's ability to generate facial expressions that are more realistic and understandable to the deaf community, since there are facial expressions that leverages depth to convey information.


\section{Conclusion}

The key idea of our approach is to create a meaningful latent space of non-manual gestures and sample from it using features related to the role of facial expressions in sign language. We propose using an architecture based on graph convolutions and upsampling operations to explore the geometric properties embedded in facial expressions at different scales. The results show that the proposed method achieved superior quality compared to several competitors and presents a high correlation with the emotions extracted from the text in the generation of facial expressions, as shown in the qualitative results. Our user study results also suggest that oral and sign language speakers are able to correctly recognize the emotions of the synthesized speaker.

\paragraph{\bf Acknowledgements} The authors thank CAPES, CNPq, FAPEMIG, and FINEP for funding different parts of this work.

{\small
\bibliographystyle{ieee_fullname}
\bibliography{main}

\begin{thebibliography}{10}\itemsep=-1pt

\bibitem{bojanowski2018optimizing}
Piotr Bojanowski, Armand Joulin, David Lopez-Paz, and Arthur Szlam.
\newblock Optimizing the latent space of generative networks.
\newblock In {\em ICML}, 2018.

\bibitem{phoenix}
N.~C. {Camgoz}, S. {Hadfield}, O. {Koller}, H. {Ney}, and R. {Bowden}.
\newblock Neural sign language translation.
\newblock In {\em 2018 IEEE/CVF Conference on Computer Vision and Pattern Recognition}, pages 7784--7793, 2018.

\bibitem{openpose}
Z. {Cao}, G. {Hidalgo Martinez}, T. {Simon}, S. {Wei}, and Y.~A. {Sheikh}.
\newblock Openpose: Realtime multi-person 2d pose estimation using part affinity fields.
\newblock {\em IEEE Transactions on Pattern Analysis and Machine Intelligence}, 2019.

\bibitem{oneeurofilter}
G\'{e}ry Casiez, Nicolas Roussel, and Daniel Vogel.
\newblock 1 € filter: A simple speed-based low-pass filter for noisy input in interactive systems.
\newblock In {\em Proceedings of the SIGCHI Conference on Human Factors in Computing Systems}, CHI '12, page 2527–2530, New York, NY, USA, 2012. Association for Computing Machinery.

\bibitem{lip_movements}
Lele Chen, Zhiheng Li, Ross~K. Maddox, Zhiyao Duan, and Chenliang Xu.
\newblock Lip movements generation at a glance.
\newblock {\em ArXiv}, abs/1803.10404, 2018.

\bibitem{enc_dec2}
Kyunghyun Cho, Bart van Merrienboer, Dzmitry Bahdanau, and Yoshua Bengio.
\newblock On the properties of neural machine translation: Encoder–decoder approaches.
\newblock In {\em SSST@EMNLP}, 2014.

\bibitem{cudeiro}
Daniel Cudeiro, Timo Bolkart, Cassidy Laidlaw, Anurag Ranjan, and Michael~J. Black.
\newblock Capture, learning, and synthesis of 3d speaking styles.
\newblock {\em 2019 IEEE/CVF Conference on Computer Vision and Pattern Recognition (CVPR)}, pages 10093--10103, 2019.

\bibitem{Duarte_CVPR2021}
Amanda Duarte, Shruti Palaskar, Lucas Ventura, Deepti Ghadiyaram, Kenneth DeHaan, Florian Metze, Jordi Torres, and Xavier Giro-i Nieto.
\newblock {How2Sign: A Large-scale Multimodal Dataset for Continuous American Sign Language}.
\newblock In {\em Conference on Computer Vision and Pattern Recognition (CVPR)}, 2021.

\bibitem{eskimez}
Sefik~Emre Eskimez, Ross~K. Maddox, Chenliang Xu, and Zhiyao Duan.
\newblock Noise-resilient training method for face landmark generation from speech.
\newblock {\em IEEE/ACM Transactions on Audio, Speech, and Language Processing}, 28:27--38, 2020.

\bibitem{FERREIRA202111}
João~P. Ferreira, Thiago~M. Coutinho, Thiago~L. Gomes, José~F. Neto, Rafael Azevedo, Renato Martins, and Erickson~R. Nascimento.
\newblock Learning to dance: A graph convolutional adversarial network to generate realistic dance motions from audio.
\newblock {\em Computers \& Graphics}, 94:11--21, 2021.

\bibitem{ginosar}
Shiry Ginosar, Amir Bar, Gefen Kohavi, Caroline Chan, Andrew Owens, and Jitendra Malik.
\newblock Learning individual styles of conversational gesture.
\newblock {\em 2019 IEEE/CVF Conference on Computer Vision and Pattern Recognition (CVPR)}, pages 3492--3501, 2019.

\bibitem{Glauert2006VANESSAA}
John R.~W. Glauert, Ralph Elliott, Stephen~J. Cox, Judy Tryggvason, and Mary Christine~Anne Sheard.
\newblock Vanessa - a system for communication between deaf and hearing people.
\newblock {\em Technology and Disability}, 18:207--216, 2006.

\bibitem{gan}
Ian~J. Goodfellow, Jean Pouget-Abadie, Mehdi Mirza, Bing Xu, David Warde-Farley, Sherjil Ozair, Aaron Courville, and Yoshua Bengio.
\newblock Generative adversarial nets.
\newblock In {\em Proceedings of the 27th International Conference on Neural Information Processing Systems - Volume 2}, NIPS'14, page 2672–2680, Cambridge, MA, USA, 2014. MIT Press.

\bibitem{greenwood18}
David Greenwood, Iain Matthews, and Stephen~D. Laycock.
\newblock Joint learning of facial expression and head pose from speech.
\newblock In {\em INTERSPEECH}, 2018.

\bibitem{FID}
Martin Heusel, Hubert Ramsauer, Thomas Unterthiner, Bernhard Nessler, and Sepp Hochreiter.
\newblock Gans trained by a two time-scale update rule converge to a local nash equilibrium.
\newblock In {\em Proceedings of the 31st International Conference on Neural Information Processing Systems}, NIPS'17, page 6629–6640, Red Hook, NY, USA, 2017. Curran Associates Inc.

\bibitem{Huang2021TowardsFA}
Wencan Huang, Wenwen Pan, Zhou Zhao, and Qi Tian.
\newblock Towards fast and high-quality sign language production.
\newblock {\em Proceedings of the 29th ACM International Conference on Multimedia}, 2021.

\bibitem{hwang2021non}
Euijun Hwang, Jung-Ho Kim, and Jong-Cheol Park.
\newblock Non-autoregressive sign language production with gaussian space.
\newblock In {\em The 32nd British Machine Vision Conference (BMVC 21)}, 2021.

\bibitem{hwang2022nonautoregressive}
Eui~Jun Hwang, Jung~Ho Kim, Suk~Min Cho, and Jong~C. Park.
\newblock Non-autoregressive sign language production via knowledge distillation, 2022.

\bibitem{you_said_that}
Amir Jamaludin, Joon~Son Chung, and Andrew Zisserman.
\newblock You said that?: Synthesising talking faces from audio.
\newblock {\em Int. J. Comput. Vis.}, 127(11-12):1767--1779, 2019.

\bibitem{eamm}
Xinya Ji, Hang Zhou, Kaisiyuan Wang, Qianyi Wu, Wayne Wu, Feng Xu, and Xun Cao.
\newblock Eamm: One-shot emotional talking face via audio-based emotion-aware motion model.
\newblock In {\em ACM SIGGRAPH 2022 Conference Proceedings}, SIGGRAPH '22, 2022.

\bibitem{KARPOUZIS200754}
K. Karpouzis, G. Caridakis, S.-E. Fotinea, and E. Efthimiou.
\newblock Educational resources and implementation of a greek sign language synthesis architecture.
\newblock {\em Computers \& Education}, 49(1):54--74, 2007.
\newblock Web3D Technologies in Learning, Education and Training.

\bibitem{Kayahan_statistical}
Dilek Kayahan and Tunga G{\"u}ng{\"o}r.
\newblock A hybrid translation system from turkish spoken language to turkish sign language.
\newblock {\em 2019 IEEE International Symposium on INnovations in Intelligent SysTems and Applications (INISTA)}, pages 1--6, 2019.

\bibitem{VAE}
Diederik~P. Kingma and Max Welling.
\newblock Auto-encoding variational bayes.
\newblock In Yoshua Bengio and Yann LeCun, editors, {\em 2nd International Conference on Learning Representations, {ICLR} 2014, Banff, AB, Canada, April 14-16, 2014, Conference Track Proceedings}, 2014.

\bibitem{koller}
Oscar Koller, Necati~Cihan Camgoz, Hermann Ney, and Richard Bowden.
\newblock Weakly supervised learning with multi-stream cnn-lstm-hmms to discover sequential parallelism in sign language videos.
\newblock {\em IEEE Transactions on Pattern Analysis and Machine Intelligence}, 42(9):2306--2320, 2020.

\bibitem{Kouremenos_statistical}
Dimitris Kouremenos, Klimis~S. Ntalianis, Georgios Siolas, and Andreas Stafylopatis.
\newblock Statistical machine translation for greek to greek sign language using parallel corpora produced via rule-based machine translation.
\newblock In {\em CIMA@ICTAI}, 2018.

\bibitem{error_propagation}
Bin Li, Jian Tian, Zhongfei Zhang, Hailin Feng, and Xi Li.
\newblock Multitask non-autoregressive model for human motion prediction.
\newblock {\em IEEE Transactions on Image Processing}, 30:2562--2574, 2021.

\bibitem{masuku}
Khetsiwe~P Masuku, Nomfundo Moroe, and Danielle van~der Merwe.
\newblock 'the world is not only for hearing people - it's for all people': The experiences of women who are deaf or hard of hearing in accessing healthcare services in johannesburg, south africa.
\newblock {\em Afr J Disabil}, 10:800, Jul 20 2021.
\newblock PMID: 34395202.

\bibitem{avatar_mcdonald}
John Mcdonald, Rosalee Wolfe, Jerry Schnepp, Julie Hochgesang, Diana~Gorman Jamrozik, Marie Stumbo, Larwan Berke, Melissa Bialek, and Farah Thomas.
\newblock An automated technique for real-time production of lifelike animations of american sign language.
\newblock {\em Univers. Access Inf. Soc.}, 15(4):551–566, nov 2016.

\bibitem{chisquared}
Mary~L McHugh.
\newblock The chi-square test of independence.
\newblock {\em Biochem Med (Zagreb)}, 23(2):143--9, 2013.
\newblock PMID: 23894860.

\bibitem{mckee}
Michael McKee, Christa Moran, and Philip Zazove.
\newblock Overcoming additional barriers to care for deaf and hard of hearing patients during covid-19.
\newblock {\em JAMA Otolaryngol Head Neck Surg}, 146(9):781--782, Sep 1 2020.
\newblock PMID: 32692807.

\bibitem{SemEval2018Task1}
Saif~M. Mohammad, Felipe Bravo-Marquez, Mohammad Salameh, and Svetlana Kiritchenko.
\newblock Semeval-2018 {T}ask 1: {A}ffect in tweets.
\newblock In {\em Proceedings of International Workshop on Semantic Evaluation (SemEval-2018)}, New Orleans, LA, USA, 2018.

\bibitem{nonmanuals}
Roland Pfau and Josep Quer.
\newblock Nonmanuals: Their prosodic and grammatical roles.
\newblock {\em Sign Languages}, pages 381--402, 01 2010.

\bibitem{lip_sync_expert}
K~R Prajwal, Rudrabha Mukhopadhyay, Vinay~P. Namboodiri, and C.V. Jawahar.
\newblock {\em A Lip Sync Expert Is All You Need for Speech to Lip Generation In the Wild}, page 484–492.
\newblock Association for Computing Machinery, New York, NY, USA, 2020.

\bibitem{rajal2}
E. Rajalakshmi, R. Elakkiya, Alexey~L. Prikhodko, M.~G. Grif, Maxim~A. Bakaev, Jatinderkumar~R. Saini, Ketan Kotecha, and V. Subramaniyaswamy.
\newblock Static and dynamic isolated indian and russian sign language recognition with spatial and temporal feature detection using hybrid neural network.
\newblock {\em ACM Trans. Asian Low-Resour. Lang. Inf. Process.}, 22(1), nov 2022.

\bibitem{rajal}
E. Rajalakshmi, R. Elakkiya, V. Subramaniyaswamy, L.~Prikhodko Alexey, Grif Mikhail, Maxim Bakaev, Ketan Kotecha, Lubna~Abdelkareim Gabralla, and Ajith Abraham.
\newblock Multi-semantic discriminative feature learning for sign gesture recognition using hybrid deep neural architecture.
\newblock {\em IEEE Access}, 11:2226--2238, 2023.

\bibitem{reimers-2019-sentence-bert}
Nils Reimers and Iryna Gurevych.
\newblock Sentence-bert: Sentence embeddings using siamese bert-networks.
\newblock In {\em Proceedings of the 2019 Conference on Empirical Methods in Natural Language Processing}. Association for Computational Linguistics, 11 2019.

\bibitem{everybodysign}
Ben Saunders, Necati~Cihan Camgoz, and Richard Bowden.
\newblock Everybody sign now: Translating spoken language to photo realistic sign language video, 2020.

\bibitem{saunders2020progressive}
Ben Saunders, Necati~Cihan Camgoz, and Richard Bowden.
\newblock {Progressive Transformers for End-to-End Sign Language Production}.
\newblock In {\em Proceedings of the European Conference on Computer Vision (ECCV)}, 2020.

\bibitem{talking_face}
Yang Song, Jingwen Zhu, Dawei Li, Xiaolong Wang, and Hairong Qi.
\newblock Talking face generation by conditional recurrent adversarial network.
\newblock In {\em IJCAI}, 2019.

\bibitem{stoll2018}
Stephanie Stoll, Necati~Cihan Camg{\"o}z, Simon Hadfield, and R. Bowden.
\newblock Sign language production using neural machine translation and generative adversarial networks.
\newblock In {\em BMVC}, 2018.

\bibitem{enc_dec1}
Ilya Sutskever, Oriol Vinyals, and Quoc~V. Le.
\newblock Sequence to sequence learning with neural networks.
\newblock In {\em Proceedings of the 27th International Conference on Neural Information Processing Systems - Volume 2}, NIPS'14, page 3104–3112, Cambridge, MA, USA, 2014. MIT Press.

\bibitem{thies}
Justus Thies, Mohamed Elgharib, Ayush Tewari, Christian Theobalt, and Matthias Nie{\ss}ner.
\newblock Neural voice puppetry: Audio-driven facial reenactment.
\newblock In Andrea Vedaldi, Horst Bischof, Thomas Brox, and Jan{-}Michael Frahm, editors, {\em Computer Vision - {ECCV} 2020 - 16th European Conference, Glasgow, UK, August 23-28, 2020, Proceedings, Part {XVI}}, volume 12361 of {\em Lecture Notes in Computer Science}, pages 716--731. Springer, 2020.

\bibitem{unterthiner2018towards}
Thomas Unterthiner, Sjoerd van Steenkiste, Karol Kurach, Raphael Marinier, Marcin Michalski, and Sylvain Gelly.
\newblock Towards accurate generative models of video: A new metric \& challenges.
\newblock {\em arXiv preprint arXiv:1812.01717}, 2018.

\bibitem{transformers}
Ashish Vaswani, Noam Shazeer, Niki Parmar, Jakob Uszkoreit, Llion Jones, Aidan~N. Gomez, undefinedukasz Kaiser, and Illia Polosukhin.
\newblock Attention is all you need.
\newblock In {\em Proceedings of the 31st International Conference on Neural Information Processing Systems}, NIPS'17, page 6000–6010, Red Hook, NY, USA, 2017. Curran Associates Inc.

\bibitem{frontalization}
Vassilios Vonikakis and Stefan Winkler.
\newblock Identity-invariant facial landmark frontalization for facial expression analysis.
\newblock In {\em 2020 IEEE International Conference on Image Processing (ICIP)}, pages 2281--2285, 2020.

\bibitem{imitating}
Haozhe Wu, Jia Jia, Haoyu Wang, Yishun Dou, Chao Duan, and Qingshan Deng.
\newblock {\em Imitating Arbitrary Talking Style for Realistic Audio-Driven Talking Face Synthesis}, page 1478–1486.
\newblock Association for Computing Machinery, New York, NY, USA, 2021.

\bibitem{FLAME}
Yao Wu and Martin Ester.
\newblock Flame: A probabilistic model combining aspect based opinion mining and collaborative filtering.
\newblock In Xueqi Cheng, Hang Li, Evgeniy Gabrilovich, and Jie Tang, editors, {\em WSDM}, pages 199--208. ACM, 2015.

\bibitem{xin2021evagcn}
Miao Xin, Shentong Mo, and Yuanze Lin.
\newblock Eva-gcn: Head pose estimation based on graph convolutional networks.
\newblock In {\em Proceedings of the IEEE/CVF Conference on Computer Vision and Pattern Recognition Workshops}, 2021.

\bibitem{yan2019convolutional}
Sijie Yan, Zhizhong Li, Yuanjun Xiong, Huahan Yan, and Dahua Lin.
\newblock Convolutional sequence generation for skeleton-based action synthesis.
\newblock In {\em International Conference on Computer Vision (ICCV)}, pages 4394--4402, 2019.

\bibitem{yan2018spatial}
Sijie Yan, Yuanjun Xiong, and Dahua Lin.
\newblock Spatial temporal graph convolutional networks for skeleton-based action recognition.
\newblock In {\em Thirty-second AAAI Conference on Artificial Intelligence (AAAI)}, 2018.

\bibitem{yin2020sentibert}
Da Yin, Tao Meng, and Kai-Wei Chang.
\newblock {SentiBERT}: A transferable transformer-based architecture for compositional sentiment semantics.
\newblock In {\em Proceedings of the 58th Conference of the Association for Computational Linguistics, {ACL} 2020, Seattle, USA}, 2020.

\bibitem{Yoon2020Speech}
Youngwoo Yoon, Bok Cha, Joo-Haeng Lee, Minsu Jang, Jaeyeon Lee, Jaehong Kim, and Geehyuk Lee.
\newblock Speech gesture generation from the trimodal context of text, audio, and speaker identity.
\newblock {\em ACM Transactions on Graphics}, 39(6), 2020.

\bibitem{controllable_facial_synth}
Hang Zhou, Yasheng Sun, Wayne Wu, Chen~Change Loy, Xiaogang Wang, and Ziwei Liu.
\newblock Pose-controllable talking face generation by implicitly modularized audio-visual representation.
\newblock {\em 2021 IEEE/CVF Conference on Computer Vision and Pattern Recognition (CVPR)}, pages 4174--4184, 2021.

\bibitem{Yang:2020:MakeItTalk}
Yang Zhou, Xintong Han, Eli Shechtman, Jose Echevarria, Evangelos Kalogerakis, and Dingzeyu Li.
\newblock Makeittalk: Speaker-aware talking-head animation.
\newblock {\em ACM Transactions on Graphics}, 39(6), 2020.

\end{thebibliography}
}

\end{document}